\documentclass[a4paper]{article}

\usepackage[english]{babel}
\usepackage[utf8]{inputenc}
\usepackage[T1]{fontenc}

\usepackage{listings}

\usepackage{lmodern}
\usepackage[autostyle]{csquotes}

\usepackage[a4paper,top=3cm,bottom=2cm,left=3cm,right=3cm,marginparwidth=1.75cm]{geometry}

\usepackage{amsmath}
\usepackage{graphicx}
\usepackage[colorinlistoftodos]{todonotes}
\usepackage{authblk}
\usepackage{rotating}
\usepackage{csvsimple}
\usepackage{float}

\usepackage{rotating}

\title{\huge Trust and Medical AI: The challenges we face and the expertise needed to overcome them}
\author[1*]{Thomas P. Quinn}
\author[1]{Manisha Senadeera}
\author[1]{Stephan Jacobs}
\author[2]{Simon Coghlan}
\author[1]{Vuong Le}

\affil[1]{\footnotesize Applied Artificial Intelligence Institute, Deakin University, Geelong, Australia}
\affil[2]{\footnotesize Centre for AI and Digital Ethics, School of Computing and Information Systems, University of Melbourne, Melbourne, Australia

* \textit{contacttomquinn@gmail.com}
}
\date{}

\Affilfont{\fontsize{4}{4}}

\begin{document}
\maketitle

\begin{abstract}

Artificial intelligence (AI) is increasingly of tremendous interest in the medical field. However, failures of medical AI could have serious consequences for both clinical outcomes and the patient experience. These consequences could erode public trust in AI, which could in turn undermine trust in our healthcare institutions. This article makes two contributions. First, it describes the major conceptual, technical, and humanistic challenges in medical AI. Second, it proposes a solution that hinges on the education and accreditation of new expert groups who specialize in the development, verification, and operation of medical AI technologies. These groups will be required to maintain trust in our healthcare institutions.

\end{abstract}

\maketitle

\section{Trust and Medical AI}

Trust underpins successful healthcare systems \cite{pellegrino_ethics_1991}. Artificial Intelligence (AI) both promises great benefits and poses new risks for medicine. Failures in medical AI could erode public trust in healthcare \cite{powles_google_2017}. Such failure could occur in many ways. For example, bias in AI can deliver erroneous medical evaluations \cite{howard_ugly_2018}, while deliberate ``adversarial'' attacks could undermine AI unless detected by explicit algorithmic defenses \cite{ma_understanding_2020}. AI also magnifies existing cyber-security risks, potentially threatening patient privacy and confidentiality.

Successful design and implementation of AI will therefore require strong governance and administrative mechanisms \cite{reddy_governance_2020}. Satisfactory governance of new AI systems should span the period from design and implementation through to re-purposing and retirement \cite{reddy_governance_2020}. In 2019, McKinsey \& Company reviewed changes needed to manage algorithmic risk in the banking sector \cite{babel_derisking_2019}. Its advice hinges on AI’s sheer complexity: just as the development of an algorithm requires deep technical knowledge about machine learning, so too does the mitigation of its risks. McKinsey \& Company discuss the need to involve three expert groups: (1) the group developing the algorithm, (2) a group of validators, and (3) the operational staff.

\begin{figure}[H]
\centering
\scalebox{1}{
\includegraphics[width=(\textwidth)]{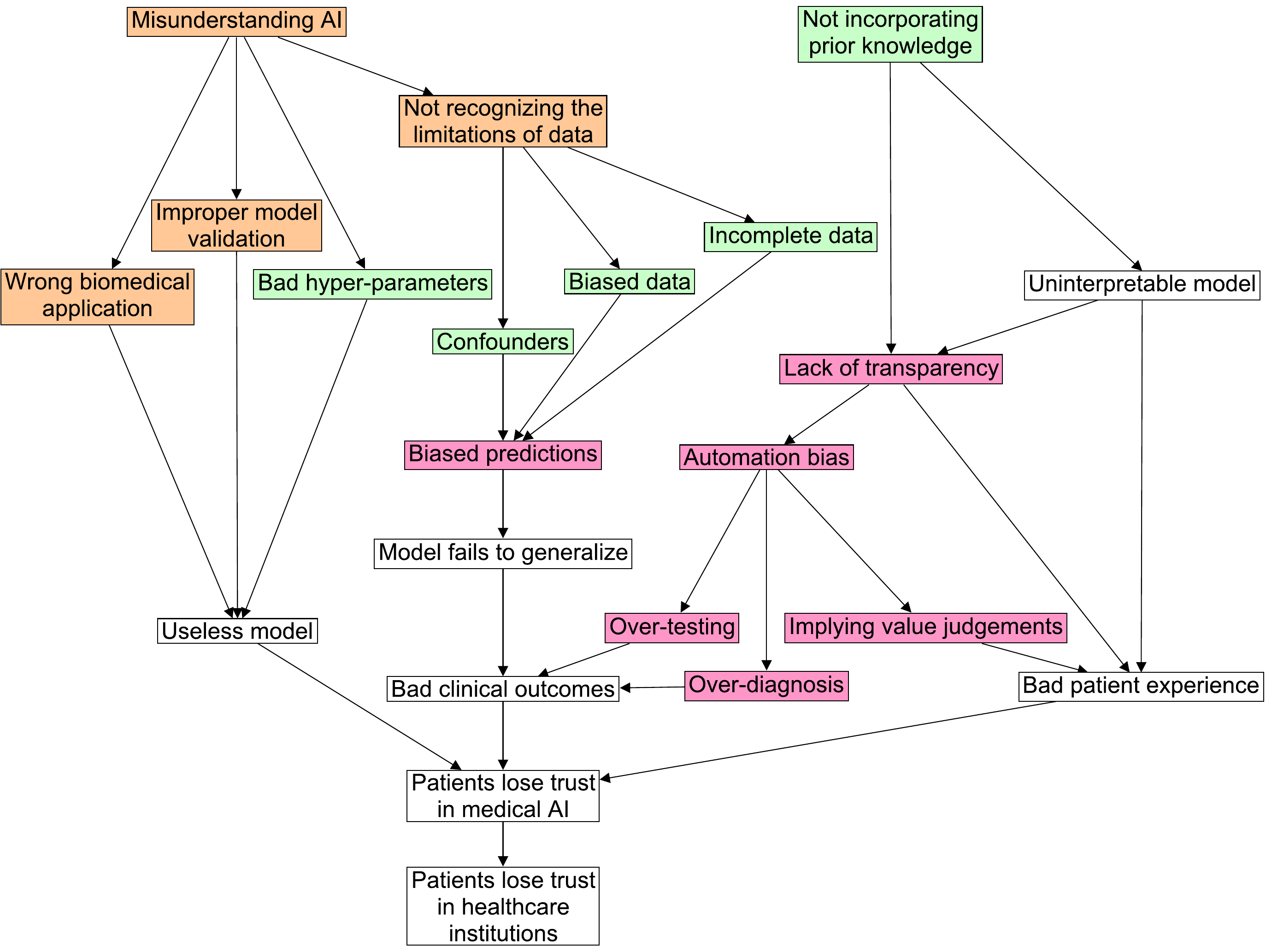}}
\caption{This figure shows how key challenges in medical AI relate to one another and to clinical care. If these challenges go unaddressed, the consequences could act concertedly to erode trust in medical AI, which could further undermine trust in our healthcare institutions. Node colour represents the type of challenge: conceptual (orange), technical (green), or humanistic (pink). Uncoloured nodes represent consequences.}
\label{fig:dag}
\end{figure}

These groups are also needed in the healthcare sector to overcome the following three key challenges in AI: (1) \textit{conceptual challenges} in formulating a problem that AI can solve, (2) \textit{technical challenges} in implementing an AI solution, and (3) \textit{humanistic challenges} regarding the social and ethical implications of AI.
This article offers concise descriptions of these challenges, and discusses how to ready expert groups to overcome them.
Recognizing these challenges and readying these experts will put the medical profession in a good position to adapt to the changing technological landscape and safely translate AI into healthcare. Conversely, failure to address these challenges could erode public trust in medical AI, which could in turn undermine trust in healthcare institutions themselves (see Figure~\ref{fig:dag}). 

\section{The Challenges We Face}

\subsection{Conceptual challenges}

Before we can translate AI into the healthcare setting, we must first identify a problem that AI can solve given the data available. This requires a clear conceptual understanding of both AI and medical practice. Conceptual confusion about AI's capabilities could undermine its deployment. Currently, AI systems cannot reason as human physicians can. Unlike physicians, AI cannot draw upon ``common sense'' or ``clinical intuition''. Rather, machine learning (the most popular type of AI) resembles a signal translator in which the translation rule is learned directly from raw data. Nevertheless, machine learning can be powerful. For example, a machine could learn how to translate a patient's entire medical record into a single number that represents a likely diagnosis, or image pixels into the coordinates of a tissue pathology. The nascent field of ``machine reasoning'' may one day yield models that connect multiple pieces of information together with a larger body of knowledge \cite{bottou2014machine}. However, such reasoning engines are currently far from practically usable.

Any study involving AI should begin with a clear research question and a falsifiable hypothesis. This hypothesis should state the AI architecture, the available training data, and the intended purpose of the model. For example, a researcher might implicitly hypothesize that a Long Short-Term Memory (LSTM) neural network trained on audio recordings of coughs from hospitalized pneumonia patients could be used as a pneumonia screening tool. Stating the hypothesis explicitly can reveal subtle oversights in the study design. Here, the researcher wants an AI model that diagnoses pneumonia in the community, but has only trained the model on patients \textit{admitted} for pneumonia. This model may therefore miss cases of mild pneumonia, and thus fail in its role as a general screening tool.
Meanwhile, model verification requires familiarity with abstract concepts like over-fitting and data leakage. Without this understanding, an analyst could draw incorrect conclusions (notably, to conclude that a model \textit{does} work when it \textit{does not}).

It is equally important to conceptualize the nature of the medical problem correctly. Although analysts might intend for a model to produce reliable results that match the standards set by human experts, this is impossible for problems in which no standard exists (e.g., because experts disagree about the pathophysiology or nosology of a clinical presentation). Even when a standard does exist, models can still recapitulate errors or biases within the training data. 

\subsection{Technical challenges}

AI is a dynamic and evolving field, and (like medicine itself) could be considered as much art as science. This makes AI much harder to use than other technologies that come with a user-manual. For example, while LSTM is widely used for sequential data, its specific application to electroencephalogram (EEG) signals requires the analyst to carefully tune dozens of so-called ``hyper-parameters'', such as the sampling rate, segment size, and number of hidden layers. These all have a major impact on performance, yet there is no universal ``rule-of-thumb'' to follow. 

AI benefits from two sources of information: (1) prior knowledge as provided by the domain expert, and (2) real-world examples as provided by the training data. With the first source of information, the model designer encodes expert knowledge into the model architecture, optimization scheme, and initial parameters, which all guide how the model learns. This is hard to do when the problem-at-hand is complex or ill-defined, as is the case in healthcare, where physician reasoning is not easily expressed as a set of concrete rules \cite{barnett_computer_1982}. 

With the second source of information, a generic model is fit to the observed data. This can deal nicely with ambiguities by discovering elaborate statistical patterns directly from the training data, and can also help update imperfect expert knowledge embedded within the algorithm. However, data-oriented models have problems too, especially when applied to healthcare, where data can be scarce or incomplete (e.g., owing to differences in disease prevalence or socio-economic factors). Such factors intensify the risk of covariate shift, confounder over-fitting, and other model biases \cite{kelly2019key}, thus reducing the trustability of purely data-oriented models.

\subsection{Humanistic challenges}

Patients are not mere biological organisms, but human beings with general and individualized needs, wishes, vulnerabilities, and values \cite{ramsey_patient_2002}. The human dimension of healthcare involves a unique professional-patient relation imbued with distinctive values and duties. This relation is widely regarded as requiring a \textit{patient-centered approach} which respects patient autonomy and promotes informed choices that align with patient values \cite{emanuel_four_1992}. Other values include the duties of privacy, confidentiality, fairness, and care, as well as the promotion of benefit (beneficence) and avoidance of harm (non-maleficence) \cite{beauchamp_principles_2001}. Medical AI must align with these values.

Many AI models are ``black-boxes'' that (for proprietary or technical reasons) cannot explain their recommendations \cite{alvarez-melis_towards_2018}. This lack of transparency could conceivably damage epistemic trust in the recommendations, and diminish autonomy by requiring patients to make choices without sufficiently understanding the relevant information \cite{grote_ethics_2020}. The use of black-boxes also makes it difficult to identify biases within models that could systematically lead to worse outcomes for under-represented or marginalized groups \cite{payrovnaziri_explainable_2020}—an important limitation given that such biases can even be present for theoretically fair models \cite{m_latent_2020}. Meanwhile, some models tend to rank treatment options from best to worst, implying value judgements about the patient’s best interests \cite{mcdougall_computer_2019}. For example, rankings could prioritize the maximization of longevity over the minimization of suffering (or \textit{vice versa}). If practitioners fail to incorporate the values and the wishes of a specific patient into their professional decisions, the AI system may paternalistically interfere with shared decision-making and informed choice \cite{elwyn_shared_2012,bjerring_artificial_2020}. A patient may even wish to follow a doctor’s opinion without any machine input \cite{ploug_right_2020}. Trust could be damaged if patients discover that healthcare workers have used AI without seeking their informed consent.

\begin{figure}[H]
\centering
\scalebox{1}{
\includegraphics[width=(\textwidth)]{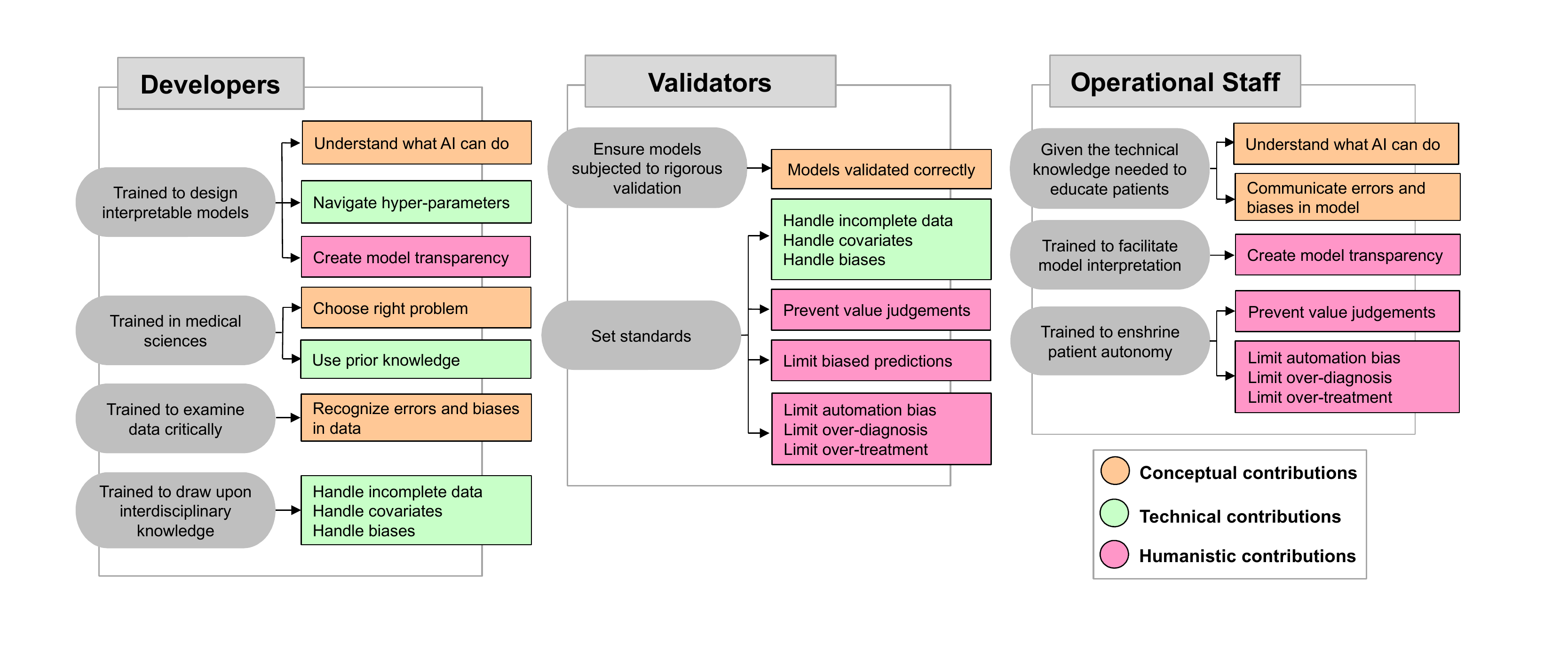}}
\caption{This table summarizes how accredited expert groups--developers, validators, and operational staff--can help overcome the key challenges in medical AI. Node colour represents the type of challenge: conceptual (orange), technical (green), or humanistic (pink).}
\label{fig:table}
\end{figure}

One problem for any powerful AI (interpretable or not) is that practitioners may come to over-rely on it, and even succumb to automation bias \cite{goddard_automation_2012}. Over-reliance, whether conscious or unconscious, can lead to harmful (maleficent) patient outcomes due to flawed health decisions, overdiagnosis, overtreatment, and defensive medicine \cite{carter_definition_2016}. 
Concern has also been raised about the incremental replacement of human beings with AI systems. For example, robot carers may soon look after older adults at home or in aged care \cite{sparrow_robots_2016}. This may deprive people of the empathetic aspects of healthcare that they want and need \cite{parks_lifting_2010}. To address the conceptual, technical, and humanistic challenges of AI in medicine, three expert groups are required: developers, validators, and operational staff (see Figure~\ref{fig:table}).

\section{The Experts We Need}

\subsection{The group developing the algorithm}

This group must understand the technical details of AI systems, but also how these details influence outcomes for patients. As such, this group should not only involve AI practitioners, but also healthcare professionals, patient advocates, and medical ethicists who together enable design processes that are flexibly sensitive to individual patient values \cite{m_latent_2020,mcdougall_computer_2019,umbrello_value-sensitive_2018}.

\subsubsection{How to ready this group:}
In the short term, we need to prioritize interdisciplinary research collaborations. Computer scientists need guidance from medical experts to choose healthcare applications that are medically important and biologically plausible. Medical experts need guidance from computer scientists to choose prediction problems that are conceptually well-formulated and technically solvable.

In the long term, we will need interdisciplinary training programs that teach computer science alongside health science, complete with accreditation through undergraduate and post-graduate degrees in \textit{digital medicine}. Both sciences rely on a precise vocabulary not readily understood by outsiders, necessitating the involvement of experts who specialize in digital medicine specifically. These degrees should also require coursework in medical ethics.

\subsection{A group of validators}

This group similarly needs to understand the technical details of AI systems in order to validate their performance in day-to-day work. Interdisciplinary collaboration will result in new knowledge production, and AI models must be constantly monitored, audited, and updated as medical knowledge advances.

\subsubsection{How to ready this group:}
In the short term, we need to apply the validation systems already available to enforce methodological rigor and safeguard patient care. 
This includes peer review, which should require that multiple disciplines critique the conceptual and technical design of AI systems, plus their humanistic implications. We should also subject AI algorithms to the same rigorous standards we apply to  evidence-based medicine \cite{evans_hierarchy_2003}—for example, by using randomized clinical trials to evaluate model performance in terms of \textit{clinical endpoints}, not just predictive accuracy.

In the long term, we will need formal institutions that are empowered to audit whether AI has been developed and deployed responsibly, giving ``AI safety'' the same scrutiny we give drug safety.
Since validation requires a strong understanding of systems-level healthcare operations, some have suggested the development of so-called ``Turing stamps'' to formally validate AI systems \cite{dalton-brown_ethics_2020}, as well as a greater involvement of official regulators like the FDA \cite{fda_proposed_2019}.

\subsection{The operational staff}

The operational staff includes any professional who works within the healthcare system.
They provide an interface between developers and validators, as well as between AI and patients.
Experience shows that computer-based recommendations may be explicitly ignored by operational staff when they find the recommendations obscure or unhelpful, with potentially disastrous consequences \cite{leveson_investigation_1993}. Operational staff can help minimize not only the risks associated with ignoring AI, but also the risks associated with over-relying on it.

\subsubsection{How to ready this group:}
In the short term, we must take staff concerns about AI safety very seriously. This includes IT staff who oversee AI systems and monitor for privacy and data security breaches. We should also encourage clinicians to use continuing medical education (CME) allowances to attend workshops and seminars on AI and AI ethics.

In the long term, we should equip healthcare workers with literacy in AI by teaching them about the conceptual, technical, and humanistic challenges as part of the professional medical curriculum. However, the intricacies of AI in medicine will additionally require opportunities for specialization. ``Digital medicine'' must become its own \textit{applied discipline}, complete with coursework and accreditation. We need ``digital doctors'' and ``digital nurses'' to calibrate patient expectations, listen and adapt to patient preferences and values, enshrine patient autonomy and decision-making capacity, and clearly communicate AI predictions alongside its limitations. We also need these experts to liaise with developers and validators in order to roll-out new technology safely.

\section{Final Remarks}

AI is a potentially powerful tool, but it comes with multiple challenges. In order to put this imperfect technology to good use, we need effective strategies and governance. This will require creating a new labor force who can develop, validate, and operate medical AI technologies. This in turn will require new programs to train and certify experts in digital medicine, including a new generation of ``digital health professionals'' who uphold AI safety in the clinical environment. Such steps will be necessary to maintain public trust in medicine through the coming AI age.

{\small
\bibliographystyle{unsrt}
\begin{bibliography}{}
\end{bibliography}
}

\end{document}